\documentclass[10pt]{article}
\usepackage[hmargin=2cm,vmargin=2cm]{geometry}
\usepackage{palatino,epsfig,latexsym,xfrac} 
\usepackage{amssymb,amsmath,amsthm,enumerate,etoolbox,paralist,url}
\usepackage{color}
\usepackage{lineno}
\usepackage{comment}
\usepackage[algo2e,ruled,vlined,linesnumbered]{algorithm2e}
\usepackage{setspace}
\DontPrintSemicolon
\usepackage[noend]{algorithmic}
\usepackage{graphicx}
\usepackage{wrapfig}
\usepackage{balance} 
\usepackage{epsfig}
\usepackage{enumitem}
\usepackage{hyperref}
\hypersetup{
    colorlinks=true,
    linkcolor=blue,
    filecolor=red,      
    urlcolor=cyan,
}
\usepackage{authblk}
\title{[Lessons Learned Report]\\ Super-Resolution for Detection Tasks in Engineering Problem-Solving}
\author{Martin Feder, Michal Horovitz, Assaf Chen, Raphael Linker and Ofer M.~Shir\footnote{Corresponding author: \href{mailto:ofers@migal.org.il}{\texttt{ofers@migal.org.il}}.}}

\begin{document}
\maketitle
\begin{abstract}
    We describe the lessons learned from targeting agricultural detection problem-solving, when subject to low resolution input maps, by means of Machine Learning-based super-resolution approaches. 
    The underlying domain is the so-called agro-detection class of problems, and the specific objective is to learn a complementary ensemble of sporadic input maps.
    While super-resolution algorithms are branded with the capacity to enhance various attractive features in generic photography, we argue that they must meet certain requirements, and more importantly, that their outcome 
    does not necessarily guarantee an improvement in engineering detection 
     problem-solving (unlike so-called aesthetics/artistic super-resolution in ImageNet-like datasets).
    By presenting specific data-driven case studies, we outline a set of limitations and recommendations for deploying super-resolution algorithms for agro-detection problems. 
    Another conclusion states that super-resolution algorithms can be used for learning missing spectral channels, and that their usage may result in some desired side-effects such as channels' synchronization. 
\end{abstract}
This technical report summarizes a research project targeting the topic of super-resolution. 
It is structured as follows: Section \ref{sec:intro} will introduce the challenge and provide necessary backgrounds. 
Section \ref{sec:approach} will present the taken approach, and particularly describe Super Resolution. 
Section \ref{sec:system} will then describe our experimental system, and Section \ref{sec:SRapplication} report on our practical observation when applying our approach.
Finally, Section \ref{sec:summary} will summarize the report and list a set of recommendations for the practitioner. 

\section{Background: Learning a complementary ensemble of sporadic input maps}\label{sec:intro}
\subsection{The challenge and motivation}
Recent developments in Artificial Intelligence (AI) combined with the advent of modern sensing technologies allow nowadays effective automated identification of detailed real-world features. Such identification has the potential to enable sophisticated machine-driven operations, e.g., autonomous vehicle control or agricultural field management, to mention a few. Indeed, the broad domain of Precision Agriculture (PA) requires accurate real-time feature detection capabilities in the field, particularly considering the recent climatic trends, the growing resiliency to pesticides, and other challenges. The core idea behind PA is the analytical consideration of the spatial variability in the field.
The practical consequences of PA translate to the farming management concept, wherein the application of specific measures are executed in order to optimize crops' input and output, i.e., maximizing crops' yield while minimizing costs and environmental impact. 
Toward this end, real-time data on soil, weather, crop maturity and stress conditions, are taken in order to measure inter and intra-field variability, and to respond accordingly. 
Data processing is usually done via a dedicated decision support system, which is designed to deal with massive amounts of data and to assist in an informed decision-making process, both for the immediate real-time short term as well as for the long term. 

A particular setup of interest in PA is concerned with spectral reflectance in the visible (VIS), near-infrared (NIR), and thermal range altogether in very high spatial resolution obtained \textit{on-demand} with unmanned aerial vehicles (UAVs)
and drones combined with satellite imagery (e.g., Landsat, Sentinel-2, and Ven$\mu$s) acquired in lower spatial resolutions, yet in a \textit{scheduled} (permanent) manner.
Although spatial resolution of the aforementioned satellites is considered low when compared to 
sensors mounted on UAVs and drones 
(3-30 meters vs.~0.5-10 cm), they provide an almost daily temporal resolution. 
At the same time, 
UAVs and drones mounted sensors 
are operated only on-demand basis. 
The inherent trade-off between spatial and temporal sensing resolutions is projected at some level onto the AI's learnability (where the focus here is on classification problems).
Scheduled satellites' sensors provide low grade information which when processed into training data is likely to enable only a ``weak'' learner and yet they constitute a persistent source of information. 
UAVs and drones mounted sensors 
provide high quality data which carries the potential to enable training into ``strong'' learners but lack the availability.
The main aim of this project is therefore to address this trade-off by composition, i.e., to be able to precisely classify prescribed states with the aid of multiple sensors, even when data are missing or being insufficient for accurate identification by independent training. Since satellite imagery is broadly accessible, success in this proposed process will save time and resources. 
In PA, for instance, it will enable cheap scans of vast areas to locate heterogeneity at the field level, and pinpoint suspected locations to be further explored by on-demand sensing.
\subsection{Novelty}
Our research suggests an AI-approach to handle multiplicity of input maps in order to accomplish successful feature identification given that learning of individual input maps is inaccurate. 
The novelty lies in the consideration of the input space as an ensemble of sporadic maps, which are to be collectively treated to enable image recognition that was heretofore viewed as either irrelevant or as too complex to address.
The core treatment idea is either to individually learn and then Bayesian-infer into a single hypothesis, or to fuse the ensemble and learn a single (aggregated) hypothesis.
We capitalized on advanced Machine Learning (ML) algorithms for implementing our proposed approaches, and to validate them on remotely-sensed multi-sensor maps within PA.
Overall, combining scheduled and on-demand maps, and utilizing scheduled sensors aided by on-demand sensors are innovative approaches for pragmatic problem-solving in PA, introducing cost efficiency and outstanding merit.
Effectively merging numerous sensors of various resolutions, bands, and costs would contribute to fully exploit each and every one of them in order to overcome challenging learning problems in PA.

\subsection{State-of-the-art}

\subsubsection*{Machine learning}
AI by means of ML is rooted in indeterminate hypotheses concerning a  problem-space. 
The process of turning a hypothesis into determinate using a solved problem-space is termed supervised training.
Successful ML training is defined as an iteratively-updated model resulting in the ability to correctly predict a solution for an unseen problem-instance.
Real-world problem-spaces, e.g., the agricultural industry, tend to be complex, due to the sheer volume of factors that need addressing.
\textit{Image recognition} is a type of important ML problems of particular interest for the agricultural industry.
In recent years, Deep Neural Networks (DNNs) have become the leading method for visual recognition, since they yield better results than
previous approaches, such as statistical methods, other ML algorithms, and image processing techniques \cite{russakovsky_imagenet_2015}. 
Prior work concerning image recognition in PA does exist, including work involving DNNs using UAVs \cite{LateBlightCNN2018,dechant_automated_2017,huang_transfer_2017,mohanty_using_2016,mahlein_recent_2012,sa_weednet_2017,sugiura_field_2016}.
The most relevant ML development with respect to the current challenge is the topic of Super-Resolution (SR), to be presented in Section \ref{sec:SR}.

\subsubsection*{Remote sensing} 
Remote sensing for crop management aims at providing spatial and spectral information for crop classification, crop condition, yield forecast, and weed/disease detection and management. Current satellite-based remotely sensed products can cover large areas, but they are limited by both their temporal (revisit times – 2 and 5 days for Ven$\mu$s and Sentinel-2 satellites, respectively) and spatial (pixel size – 5 and 10 m for Ven$\mu$s and Sentinel-2, respectively) resolutions, when compared to a UAV or a drone. 
One of satellite imaging’s challenges is dealing with pixels that have multiple objects with different spectral signatures (e.g., plants and soil). Such pixels are called mixed-pixels. 
Images acquired by UAVs and drones, which have a much  higher spatial resolution, contain many more pure (as opposed to mixed) pixels, which makes vegetation detection and differentiation much easier. Similarly, high spatial resolution allows for a precise estimation of the vegetation cover fraction. Beside their spatial and temporal resolution, remote sensing platforms differ in terms of the sensors they carry. 
To date, the measurements most commonly used in applications related to agriculture consist of passive measurements in the VIS ($\approx$ 380-740nm), NIR ($\approx$ 750nm-1.4$\mu$m) and thermal ranges ($\approx$ 8-15$\mu$m). Regardless of the spectral range, the sensor is either broadband,  multi-spectral (i.e., a small number of relatively wide bands) or hyper-spectral (i.e., a typically high number of narrow bands). 
Notably, sensing platforms carried most by satellites include only a few, relatively wide, spectral bands, that can obviously not be changed according to the users' needs. 
By comparison, UAV and drones' platforms are much more flexible and, if used on-demand for a specific need as suggested in the present work, the sensing platform carried by the UAV or the drone can be selected to best fit the monitoring needs. 
\section{Approach}\label{sec:approach}
\subsection{Super-Resolution (SR) Algorithms}\label{sec:SR}
\begin{figure}[h!]
    \centering
    ``Aesthetics-SR'': explicit role\\
 \fbox{\includegraphics[width=0.5\columnwidth]{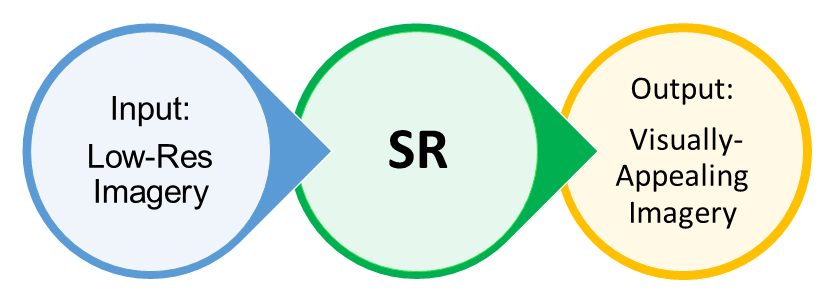} }
 \medskip
 \\
    ``Detection/Engineering-SR'': implicit role\\
 \fbox{ \includegraphics[width=0.7\columnwidth]{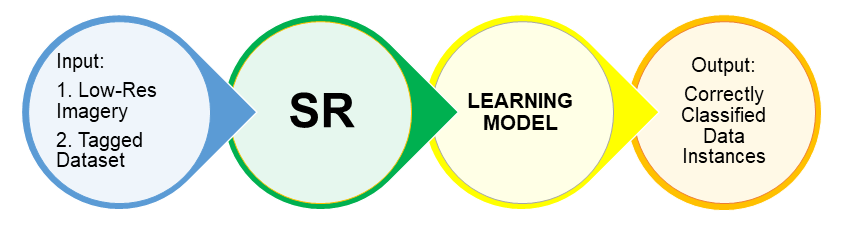}}
    \caption{Comparing the high-level problem-solving of ``Aesthetic'' versus ``Detection/Engineering'' SR tasks by means of process diagrams.\label{fig:SR}}
\end{figure}

The capacity of learning algorithms to induce high resolution imagery from lower resolution inputs, based upon pre-training, is referred to 
super-resolution (SR) \cite{shermeyer2019effects}.
Various algorithms and approaches have been devised (see, e.g., \cite{rs10081216,8710329,8909276,8920910}).
While SR algorithms are branded with the capacity to enhance various attractive features in generic photography, we argue that they must meet certain requirements when applied to engineering detection challenges (unlike aesthetics or artistic challenges).
Most importantly, problem-solving of ``Aesthetic'' versus ``Detection/Engineering'' SR-tasks differs already at the process definition, especially at the input-output levels; 
When SR is applied to an Aesthetic task, it performs the primary operation of transforming a degraded input imagery into an enhanced, visually appealing imagery -- which constitutes the output. SR is explicit then. 
However, when SR is applied within a learning pipeline its role is reduced to performing an \textit{auxiliary} operation of enhancing data instances that undergo training/testing. The output is the classification response of the learning model. SR may be considered implicit then. The comparison between the two tasks is illustrated by means of process diagrams in Figure \ref{fig:SR}.

Consequently, SR models that were successfully trained on ImageNet-like datasets are not necessarily potent for engineering detection problem-solving.
Altogether, we argue that there could not exist a generic SR algorithm with the capability to enhance any given input of an arbitrary sort (equivalently, to some degree, to the No Free Lunch set of theorems \cite{Wolpert97nofree}).
In other words, the capacity to enhance low-resolution imagery is feature-dependent, and clearly requires suitable pre-training that relies on instances of the same domain. 
Furthermore, additional requirements exist, especially concerning the available spatial resolution versus the scale of the targeted features. 
One of the primary goals of this paper is to outline a set of recommendations concerning the feasibility and infeasibility of SR usage.

\subsection{Formulation}
Given a multi-source remotely-sensed set of input maps, the primary objective is to devise an automated procedure for classification of predefined states that achieves successful learning.
All input maps are assumed to cover a specific area of interest, e.g., an agricultural field, whose targeted states are well-defined 
(for instance, a binary classification problem with a healthy crop state versus a diseased state). 
Given a classification problem, the system encompasses at most $N$ remote sensors numbered from $1$ to $N$, where the first $N_s$ sensors are scheduled sensors and the remaining $N_d$ sensors are on-demand sensors, $N=N_s+N_d$.
The underlying classification problem is assumed to be static in the sense that each individual map carries information to infer or refute the targeted states (unlike dynamic problems that require a time-series of maps to deduce the state).

Explicitly, we denote by $\mathcal{M}_k$ the set of all input maps of sensor $k$. 
Then, all input maps constitute a disjoint union of two subsets, the set of all the \textbf{scheduled maps} 
$$\mathcal{M}^{s}=\bigcup_{i=1}^{N_s} \mathcal{M}_i$$
and the set of all the \textbf{on-demands maps} 
$$\mathcal{M}^{d}=\bigcup_{i=N_s+1}^{N} \mathcal{M}_i~.$$
When no sensing takes place, the related map will be in a void state.

\subsubsection{Assumptions}
\begin{itemize}
    \item[\textbf{A1}] For practical learning purposes, each map $m \in\mathcal{M}_k$ may be segmented into an effective feature space of dimension $\mathcal{D}_k$.
    \item[\textbf{A2}] We assume that each set $\mathcal{M}_k$ is learnable by a classifier $\mathcal{C}_k$ with an expected error rate of $\epsilon_k$, when averaged over all input maps within the set (formally, this assumption translates into PAC-learnability \cite{Shalev2014Understanding} using sample size $n(\epsilon,\delta)$, with pragmatic $\epsilon,\delta$ values, distinguishable from a random learning hypothesis). Note that this learnability perspective focuses on the existence of a learning function, while assuming the existence of a potent learning algorithm.
    \item[\textbf{A3}] We assume that learning of the scheduled sets inevitably leads to high error rates $\forall i=1\ldots N_s~\epsilon_i \gg 0$ (due to their low spatial resolution). Since straightforward boosting techniques fail, on-demand sensing and the learning of the acquired sets are much needed for sufficiently accurate classification.
    \item[\textbf{A4}] The on-demand sensing is hard and expensive to perform, while the scheduled sets are easily accessible. Hence, combining these two classes of maps' sets is necessary.
\end{itemize}

\subsubsection{Concept Outline}
Collecting data along a dedicated period, e.g., a single agricultural season, may serve to construct a training set comprising all $N$ types of maps.
Tagging (labeling) of the targeted states is assumed to be applied to the dataset.
A consecutive period may become the testing phase.
The training and the testing are applied in the following manner. 
At first, the learning data is limited only to the $N_s$ ``scheduled maps'' (simulating a scenario where only scheduled sensors are available).
Upon successful training, the learner is examined at classifying the targeted states with ``scheduled maps'' as the testing data. 
If the classification output is provided with high accuracy (with respect to the tagged states of the dataset), no action is taken; else, that is below some accuracy threshold, ``on-demand maps'' are to be used, according to some selection criterion (simulating a scenario where on-demand sensors are deployed), to play a role in both the training and the testing parts.

\subsubsection{Algorithm}
The main research question is how to effectively learn such input maps $\bigcup_{k=1}^{N_s} \mathcal{M}_k$ as a classification problem given targeted states 
and by employing a sensible number of PAC-learning classifiers.

The plan is to treat this challenge by obtaining an SR-learner to the scheduled maps using the acquired on-demand maps. The obtained SR-learner is denoted as $\mathcal{S}\mathcal{R}$.
Formally, the application of $\mathcal{SR}$ on an input map $\mathcal{M}_k$ induces another map (or, possibly, a set of maps, since $\mathcal{M}_k$ is defined as a set of maps by itself) , which is denoted by $\mathcal{SR}\left[ \mathcal{M}_i\right]$. 
Altogether, the proposed pipeline is the following (Algorithm \ref{Algo:A0}), where its termination criterion is set with respect to the error rates of the low-/high-resolution and SR classifiers:\footnote{The error rates' notation is rooted in $\ell$ for low-resolution and $h$ for high-resolution, but the careful reader should not be confused by the fact that $\epsilon_h < \epsilon_{\ell}$.}
{\LinesNotNumbered
\begin{algorithm2e}[h]
\caption{Basic pipeline for obtaining $\mathcal{SR}$ per an engineering classification task \label{Algo:A0}}
\setstretch{1.2}
\KwIn{low-res \textit{scheduled} maps, $\bigcup_{i=1}^{N_s} \mathcal{M}_i$, with tagged targeted states\\
\hphantom{\textbf{Input:} }high-res \textit{on-demand} maps, $\bigcup_{i=N_s+1}^{N} \mathcal{M}_i$, with tagged targeted states\\
\hphantom{\textbf{Input:} }accuracy improvement satisfactory threshold $\varphi$
}
\KwOut{an $\mathcal{SR}$-assisted classifier $\mathcal{C}$}
\begin{enumerate}[rightmargin=24pt,leftmargin=6pt]
\vspace{3mm}
\item{Train and test a classifier $\mathcal{C}$ on the \textit{scheduled} maps, $\bigcup_{i=1}^{N_s} \mathcal{M}_i$; denote the testing error rate by $\epsilon_{\ell}$.}
\item{Train and test a classifier $\mathcal{C}$ on the \textit{on-demand} maps, $\bigcup_{i=N_s+1}^{N} \mathcal{M}_i$; denote the testing error rate by $\epsilon_h$.}
\item{Learn an SR-model, denoted by $\mathcal{S}\mathcal{R}$, as follows: train and test a model to receive \textit{scheduled} maps, $\bigcup_{i=1}^{N_s} \mathcal{M}_i$, as input and produce their \textit{on-demand} maps $\bigcup_{i=N_s+1}^{N} \mathcal{M}_i$ as the correct output.}
\item{Train and test a classifier $\mathcal{C}$ on the enhanced (scheduled) maps, $\bigcup_{i=1}^{N_s} \mathcal{SR}\left[ \mathcal{M}_i\right]$; denote the overall testing error rate by $\epsilon_{SR}$.}
\item{Assess $\epsilon_{\ell},~ \epsilon_h,~\epsilon_{SR}$: \textbf{if} $\left(\epsilon_{\ell}-\epsilon_{SR}\right)/\epsilon_{\ell} \geq \varphi$ then \textbf{return} $\mathcal{SR}$; \textbf{else}, reiterate $1\mapsto 4 $ with a different 
$\mathcal{C}$ and/or $\mathcal{SR}$ model.}
\end{enumerate}
\end{algorithm2e}
}
\section{Experimental System: Irrigation Monitoring} \label{sec:system}
Precision irrigation can be defined as matching water application to the crop needs which is rarely uniform in space, time and amount. One of the main sources of crop growth variability in semi-arid climates is due the lack of irrigation uniformity, which can be due to bad design of the irrigation system or operational failures (leaks, clogging). Awareness of the extent and severity of possible uniformity problems requires spatial and temporal crop water status information at sufficient resolution and accuracy, attainable mainly by aerial sensing and imagery 
\cite{el2011approach,haghverdi2015perspectives,zhang2011spatial}.
New enabling technologies for precision irrigation information acquisition became affordable and popularized with the appearance in the open market of low cost, high performance small-sized UAV and drones carrying high-resolution digital camera, as well as fast revisiting, low cost satellite services in the VIS (RGB), NIR and long wave (Thermal) infrared ranges.
Most irrigation systems, fixed and especially mobile, suffer from a lack of uniformity in terms of water coverage. Mapping the crop water status in space and time is crucial for variable rate irrigation (VRI) application and for adapting irrigation according to the specific crop water requirements. 
Crop thermal imaging via remote sensing technology enables mapping the state of water in the field. Spectral vegetation indices (VI) built from combinations of channels at various wavelengths allow for better information extraction from remotely sensed data because they reduce the effects of soil, view angle and topography, while enhancing the focus on the visibility of the vegetation \cite{hunt2013visible}. 
The objective of this experimental system’s research was to test the efficiency of ML models to detect irrigation variability of a lateral move and pivot irrigation machines, and to recognize malfunction in some cases.
\setlength\fboxsep{2.5pt}
\setlength\fboxrule{0.5pt}
\begin{figure}[h!]
\centering
\fbox{\includegraphics[width=0.96\linewidth]{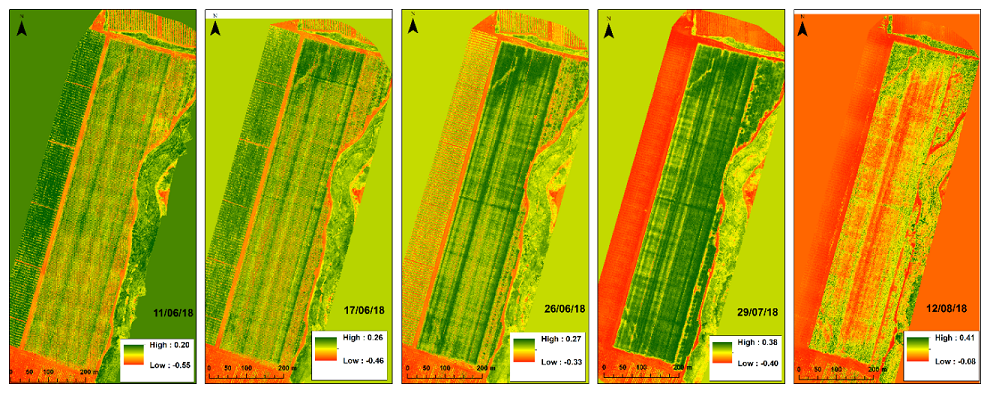}}
\caption{GRVI images of the cotton field in Hagoshrim with dates indicated next to each image. GRVI values decrease from green to yellow to red.\label{fig:1}}
\end{figure}
%
%
\begin{figure}[t]
\centering
\fbox{\includegraphics[width=0.96\linewidth]{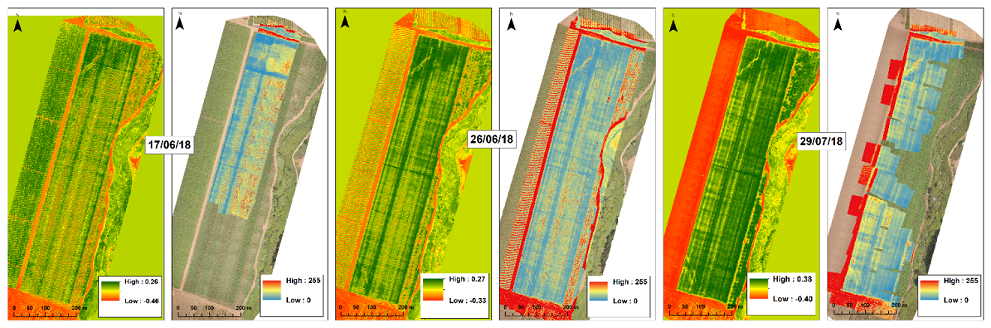}}
    \caption{Thermal (right of pair) and corresponding GRVI images (left of pair) of the cotton field in Hagoshrim with dates indicated next to each image. Thermal images: red indicating hotter temperatures, blue colder. GRVI values decrease from green to yellow to red.\label{fig:2}}
\end{figure}
%
%
\begin{figure}[h]
    \centering
 \fbox{\includegraphics[width=0.96\linewidth]{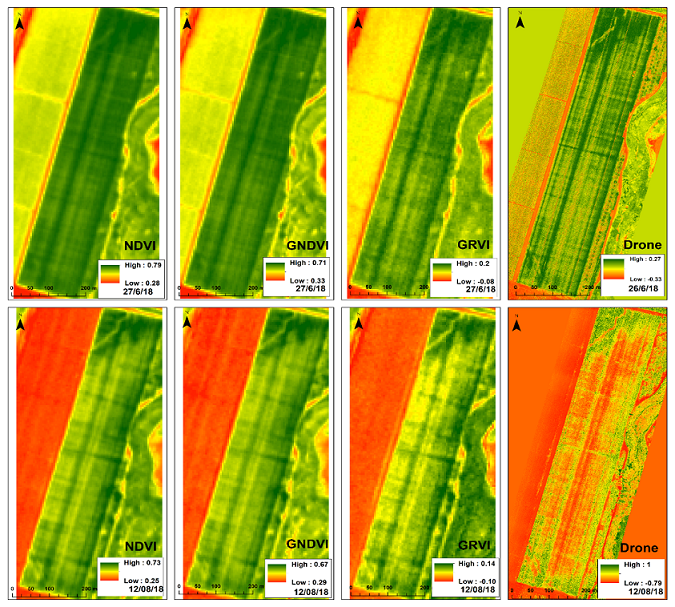}}
    \caption{NDVI, GNDVI, and GRVI vegetation indices based on Venus imagery of 27/6/2018 (top) and 12/08/2018 (bottom) and the corresponding UAV GRVI image of the cotton field at 26/6/2018 and 12/08/2018, respectively.\label{fig:3}}
\end{figure}
\begin{figure}[h]
    \centering
 \fbox{\includegraphics[width=0.96\linewidth]{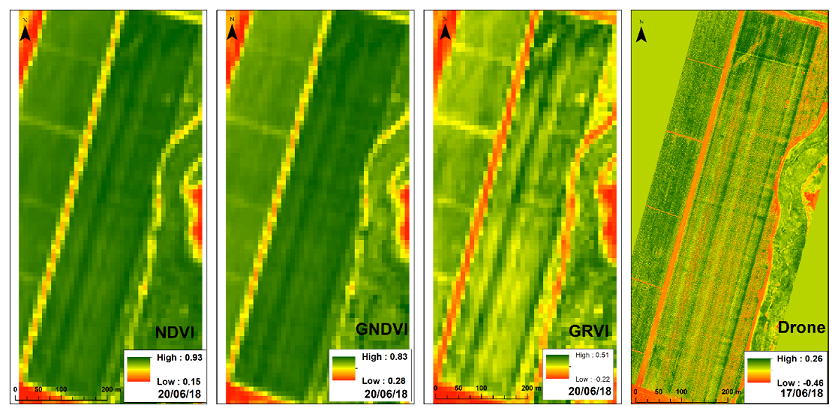}}
    \caption{NDVI, GNDVI, and GRVI vegetation indices based on a Sentinel-2 imagery of the cotton field in Hagoshrim test site from 20/06/2018 and the corresponding UAV GRVI Image from 17/06/2018.\label{fig:4}}
\end{figure} 

In what follows, we describe preliminary computational tasks that we addressed with respect to given irrigation monitoring datasets. The following section will discuss the experimental observations concerning the application of SR algorithms.

 \subsection{Insights and Limitations}
Since the Ven$\mu$s satellite resolution was believed to be insufficient for classifying irrigation features, we down-scaled images taken by drones to synthesize (``simulate'') satellite lower-resolution imagery. 
The idea was to define the resolution at which the irrigation signal can still be detected.
Since the desired substitute imagery was a mosaica of multi-spectral images, and due to the spatial misalignment across bands, there existed an initial best resolution constraint.

\subsection{Irrigation Learnability}
Here we aimed at the classification of various irrigation policies through statistical and classical machine learning methods in Neve Yaar.
Description: Use irrigation regimens to establish the separability of the irrigation areas using different sensory inputs.

\begin{figure}[h!]
    \centering
 \fbox{\includegraphics[width=0.96\linewidth]{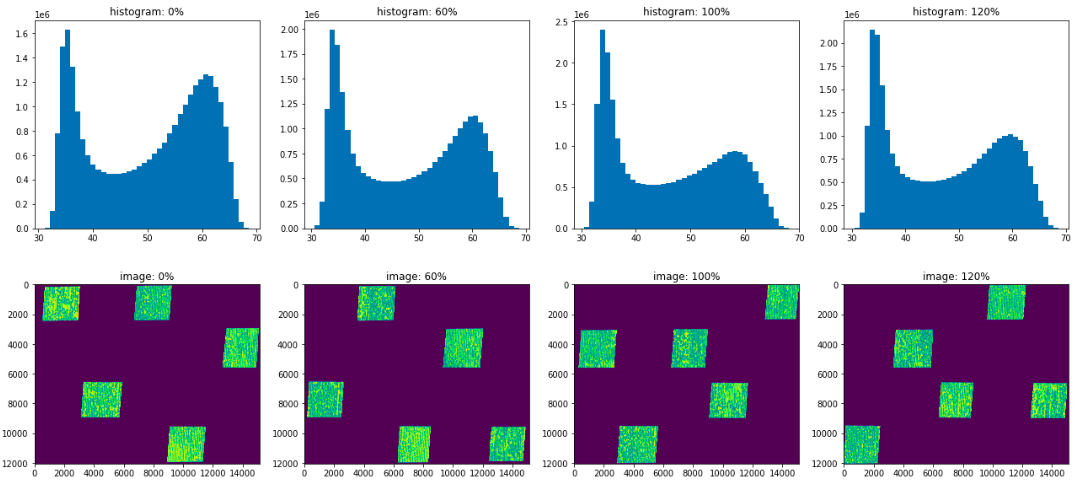}}
    \caption{Thermal Statistics of Neve Yaar by Irrigation Regime - the histograms [TOP] corrspond to the field segments [BOTTOM] that were irrigated according to one of the policies: $\left\{0\%,~60\%,~100\%,~120\% \right\} $\label{fig:statsTherm}.}
\end{figure}

Sensory image and index-based sensory were explored across the dates, to investigate whether the problem of separating the plots by irrigation regimen was learnable in each. 
Moreover, each image was downscaled in order to also estimate its learnability in lower resolutions.
The thermal statistics of the dataset is shown in Figure \ref{fig:statsTherm}, depicting the field segments on which an irrigation policy was applied (bottom row) and their associated thermal histograms (upper row).
\section{Application of SR} \label{sec:SRapplication}
In what follows, we outline the concrete SR-related tasks that we addressed with respect to given irrigation monitoring datasets.

\subsection{Baseline Deployment} \label{sec:baseSR}
Description: Use the Never-Yaar multi-spectral data as a proxy for up-scaling from the Ven$\mu$s multi-spectral data.

Since the goal was to enhance readily available Ven$\mu$s satellite imagery with on-demand high resolution mosaicas, the acquired imagery (satellite and drone mosaicas from 4 dates) was aligned against an irrigation map, georeferenced to rectify alignment imprecision, trimmed to a bounding box around the irrigation map, and resized to a base common resolution of 2.5 cm per pixel (through down-scale pooling or up-scaling by linear extrapolation), resulting in a 3D-matrix image-set.
We trained models with different input-output pairings using different channels from the same date. Since the objective was to produce a model that could enhance an image, we sought to understand the circumstances at which a model could reliably add information to an input. Super-resolution training was accomomplished by reducing the resolution of the input space for any given analysis. Super-resolution was made by downscaling by 2, 4 and 8 factor reductions of the input space. The super-resolution layout was up to 10 sequential convolution and deconvolution layers and~300 thousand parameters per layer or less. 
Results were visually examined and assessed using image similarity metrics. We started with an input space that contained all of the channels with a target of specific channels and proceeded to exclude input bands. The multi-spectral imagery was trained against RGB target.  reduced to lower resolution, and the targets were switched between RGB, thermal imagery and NDVI.

\subsection{Regime Identification}
Description: Use the Never-Yaar multispectral data to classify patches by irrigation regimen by means of SR.

The former step in \ref{sec:baseSR} was to establish a baseline SR. The added step was meant to try and classify the post-SR image (whether it was approximated to be multispectral, thermal or NDVI) with the irrigation regimen as the target.
We first attempted this both directly, through the irrigation plot labeling, but also through 

Results: (The algorithm failed to converge at the original resolution for the test split. likewise for the downscaled resolutions)

\subsection{Irrigation segments using Goshrim}
Description: Use Goshrim to test for the ability to identify localized irrigation patterns.

The Neve-Yaar plot was a planned experiment of different irrigation regimens. That said, since the plot assignment was of large areas, the difference in application would not have necessarily been locally homogeneous.
In contrast, though the Goshrim plot was not experimentally planned for differential irrigation, an irrigation malfunction resulted in irrigation patterns. Since the irrigation patterns were clearly observable and differentiable it was expected that the patterning would be more easily detectable than in the Neve Yaar dedicated irrigation plots. 
At the same time, there was no definition of irrigation policies here. 
Hence, a pseudo-irrigation binary classification problem was approximated by first passing a temperature threshold using the thermal sensory and later by manually adjusting the output map for whatever areas consistency could be assessed and correct visually.

\begin{figure}[h!]
    \centering
 \fbox{\includegraphics[width=0.6\linewidth]{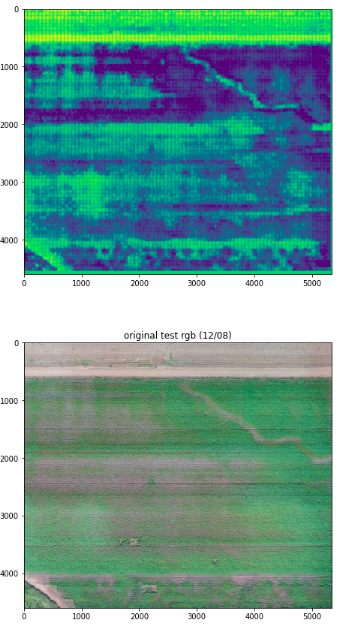}}
    \caption{Infer thermal map (top) from RGB imagery (bottom)\label{fig:predTherm}}
\end{figure}

\subsection{SR using Ven$\mu$s alone}

Description: The Venus satellite takes raw images at a resolution of ~5m. A dedicated algorithm is then applied to create an undistorted image. The algorithm relies on aligning several images from proximal dates. The alternatively algorithm requires reducing the raw image to a lower resolution.

The choice of dataset and problem were to circumvent the spatial and temporal limitations (limited plot size, time repetitions) inherent in Neve Yaar by working directly with Ven$\mu$s-satellite data. The task was to replicate the algorithm that corrects the raw image (5m), and possibly improve upon it by not requiring multiple . 
The goal was to investigate whether we can correct the original image (5m) for inference without reducing the resolution or requiring multiple images.

The result metrics suggest the model output is closer than the raw image. 
Visually, however, the model has not converged, irrespective of the metrics.

\section{Summary and Recommendations} \label{sec:summary}
The Newe Yaar dataset was not learnable at the high-resolution level -- evidently, the classification of the irrigation policies was not generalizable, which rendered the proposed pipeline (Algorithm \ref{Algo:A0}) unfit for this task. 
Moreover, an attempt to obtain data fusion of maps taken across different dates was unsuccessful due to lack of normalization. 
Following that, we conducted several computational tasks that focused on the challenge of SR learning. \\
Next, we describe our concluding checklist for the data scientist who wishes to address a Detection/Engineering SR problem.
\subsection{A Concluding Checklist}
\begin{enumerate}
    \item Assess the feature size (linked to the targeted state) versus the pixel size at the scheduled maps.
    \item Assess the resolution gaps (e.g., satellites versus drones) -- is the factor reasonable? 
    \item Sensory noise (with respect to a calibrated ``white screen''): Assess the signal-to-noise ratio
    \item Spatial noise (due to geo-referencing): Assess the registration error (minor shifts can alter tagging)
    \item Address this question: Is it possible to fuse datasets across multiple dates (with respect to normalization and/or noise)?
\end{enumerate}

\section*{Acknowledgments}
This research was supported by the Ministry of Science and Technology, Israel.\\
The authors thank B.~Yazmir for his contributions that ignited the SR line of work and for conducting a comprehensive literature survey in that direction.



\newcommand{\etalchar}[1]{$^{#1}$}
\providecommand{\bysame}{\leavevmode\hbox to3em{\hrulefill}\thinspace}
\providecommand{\MR}{\relax\ifhmode\unskip\space\fi MR }
\providecommand{\MRhref}[2]{%
  \href{http://www.ams.org/mathscinet-getitem?mr=#1}{#2}
}
\providecommand{\href}[2]{#2}

\end{document}